\documentclass[12pt,leqno,twoside,sort]{amsart}

\usepackage{amssymb,amsmath,amsthm,soul,color}
\usepackage[normalem]{ulem}
\usepackage{amssymb,amsmath,amsthm}
\usepackage{mathrsfs}
\usepackage[utf8]{inputenc}
\usepackage{todonotes}
\usepackage[left=2cm, right=2cm, top=2cm, bottom=2cm]{geometry}
\usepackage{url}
\usepackage[numbers,sort&compress]{natbib}
\usepackage{fixmath}

\usepackage[T1]{fontenc}
\usepackage{lmodern}
\usepackage{microtype}

\usepackage{xurl}

\usepackage{mathtools}

\usepackage[colorlinks=true,urlcolor=blue,
citecolor=red,linkcolor=blue,linktocpage,pdfpagelabels,pagebackref=true,
bookmarksnumbered,bookmarksopen,breaklinks=true]{hyperref}

\newenvironment{nouppercase}{%
  \renewcommand{\uppercasenonmath}[1]{}}{}

\usepackage{xcolor,cancel}

\usepackage{enumitem,graphicx}

\usepackage{caption}
\usepackage{subcaption}

\usepackage{float}

\usepackage[most]{tcolorbox}
\usepackage{fontawesome5}

\definecolor{lightblue}{RGB}{173,216,230}
\definecolor{lightgreen}{RGB}{200,255,200}
\definecolor{lightred}{RGB}{255,200,200}
\definecolor{lightgray}{RGB}{230,230,230}

\usepackage[mathlines]{lineno}

\numberwithin{equation}{section}

\usepackage{xspace}
\newcommand{\pro}{\textit{gemini-2.5-pro-03-25}\xspace}
\newcommand{\flash}{\textit{gemini-2.5-flash-04-17}\xspace}
\newcommand{\othree}{\textit{o3-mini-high}\xspace}
\newcommand{\ofour}{\textit{o4-mini-high}\xspace}
\newcommand{\rone}{\textit{r1}\xspace}
\newcommand{\sonnet}{\textit{sonnet-3.7}\xspace}

\begin{document}

\title{Putnam-like dataset summary: LLMs as mathematical competition contestants}

\author[B. Bieganowski]{Bartosz Bieganowski}
\address[B. Bieganowski and D. Strzelecki]{\newline\indent
			Faculty of Mathematics, Informatics and Mechanics, \newline\indent
			University of Warsaw, \newline\indent
			Banacha 2, 02-097 Warsaw, Poland}	
			\email{\href{mailto:bartoszb@mimuw.edu.pl}{bartoszb@mimuw.edu.pl}}	
			\email{\href{mailto:dstrzelecki@mimuw.edu.pl}{dstrzelecki@mimuw.edu.pl}}

\author[R. Skiba]{Robert Skiba}

\address[R. Skiba and M. Topolewski]{\newline\indent
			Faculty of Mathematics and Computer Science, \newline\indent
			Nicolaus Copernicus University in Toru\'n, \newline\indent
			Chopina 12/18, 87-100 Toru\'n, Poland}	
			\email{\href{mailto:robo@mat.umk.pl}{robo@mat.umk.pl}}	
			\email{\href{mailto:woland@mat.umk.pl}{woland@mat.umk.pl}}
			
\author[D. Strzelecki]{Daniel Strzelecki}
			
\author[M. Topolewski]{Mateusz Topolewski}

\date{}	\date{\today}

\begin{abstract} 
In this paper we summarize the results of the Putnam-like benchmark published by Google DeepMind. This dataset consists of 96 original problems in the spirit of the Putnam Competition and 576 solutions generated by LLMs. We analyze the performance of models on this set of problems to verify their ability to solve problems from mathematical contests. We find that top models, particularly Gemini 2.5 Pro, achieve high scores, demonstrating strong mathematical reasoning capabilities, although their performance was lower on problems from the 2024 Putnam competition. The analysis highlights distinct behavioral patterns
among models, including bimodal scoring distributions and challenges in providing fully rigorous justifications.

\end{abstract} 
\begin{nouppercase}
\maketitle
\end{nouppercase}
\pagestyle{myheadings} \markboth{\underline{B. Bieganowski, R. Skiba, D. Strzelecki, M. Topolewski}}{
		\underline{Putnam-like Dataset Summary}}

\section{Introduction}
General-purpose large language models have recently shown significant progress in mathematical reasoning. It was observed that LLMs are obtaining high scores in final-answer benchmarks (see \cite{x2}). In such benchmarks, models are asked only for the final answer to a mathematical problem, which is then compared to the correct one. Models do not produce complete, rigorous solutions (or they are not verified), but they provide only the final answer, which is usually a number. It is clear that such benchmarks are not able to capture the real mathematical capability of the model, except in the case of extremely challenging problems, see \cite{x7}.

Hence, there is a need to analyze the behavior of LLMs on mathematical proof-like problems, in particular, where the aim of the problem is to prove or disprove some property. Such problems are quite popular in mathematical competitions. The behavior of LLMs on proof-based problems was studied on problems from a number of  mathematical competitions, including the 2025 USA Mathematical Olympiad \cite{x1,x4}, the International Mathematical Olympiad \cite{x1, x5} (also formalized in Lean \cite{x6}), and the International Mathematics Competition for University Students (IMC) \cite{x1,x8}. See also \cite{x9,x12} for a wide benchmarks of LLM-generated mathematical proofs for the problems originated in mathematical competitions. The parallel topic is the benchmarking of LMMs on research-level problems that requires a rigorous proof, see \cite{x10}.

On September 23, 2025, Google DeepMind released the dataset, called Putnam-like (\url{https://github.com/google-deepmind/eval_hub/tree/master/eval_hub/putnam_like}), with 96 mathematical problems on the level of competitions for students like the William Lowell Putnam Mathematical Competition (\url{https://maa.org/putnam/}) or IMC (\url{https://www.imc-math.org.uk/}). These problems were prepared in cooperation with Google DeepMind by the authors of this article. 

The dataset consists of 8 sets of problems that mimic the style of the Putnam competition, i.e., in each of these sets there are two parts (A and B) with six problems of increasing difficulty.  The data published by Google DeepMind has a rich structure. For each problem there is given a proposed solution (rubrics) divided into steps worth some points, summing up to 10. Six AI models were tested on each problem, including Google's Gemini family (\emph{gemini-2.5-pro-03-25} and \emph{gemini-2.5-flash-04-17}), models of OpenAI (\emph{o3-mini-high} and \emph{o4-mini-high}), Anthropic's Claude Sonnet (\emph{sonnet-3.7}) and the first generation of DeepSeek model (\emph{r1}). The models' solutions were verified and graded by human experts (this article's authors) and by the Google DeepMind's \emph{gemini-2.5-pro\_20250718} model. All the grades, along with the comments, are published in the dataset. 

In this paper we summarize the LLMs' ability to solve the problems in natural language proposed in the dataset. These problems require different mathematical skills to be solved, and the proposed solution has to be completely, fully rigorously proven, even if some numerical answer can be provided. Note that, as described in the dataset's readme file, the models were prompted only with the problem statement, without any requirements for the quality of solutions. However, in the grading process performed by experts, the solutions were graded as being given during the mathematical contests, especially: the usage of a very precise numerical approximations was prohibited (as impossible to be done during a competition), the references for the non-standard theories were required, and all steps of the reasoning were expected to be written.

In our study we analyze six levels of problems, i.e., we don't distinguish between part A and part B of the Putnam competition structure. Moreover, to each problem in the dataset we have assigned one of the categories: linear algebra, abstract algebra, analysis (including inequalities), discrete mathematics, probability, number theory, and polynomials. Since the Putnam-like dataset is expected to provide Lean formalization for each problem, there are no problems in geometry or combinatorics in this dataset. Lean formalizations are provided in another repository (\url{https://github.com/google-deepmind/formal-putnam-like}). 

In this article, we focus only on the grades given by human experts, treating them as the most accurate. The correlation of the human and automatic grades needs to be studied in future work.

This study is another validation of the LLMs on Putnam-style problems.  The study of LLMs solutions to the Putnam Competition problems given in natural language was done in the recent Putnam-AXIOM benchmark \cite{x2}. Note that this dataset collects problems from the actual Putnam competition, possibly with minor symbolic modifications, so most of these problems were contained in the LLMs' training data. The Putnam-like dataset is not affected by this type of data contamination.

There is also a very interesting study of the AI models solving reformulations of Putnam Competition problems into formal languages like Lean, Isabelle, or Coq \cite{x3}.  

The goal of this paper is to investigate and compare the behaviors of different language models on Putnam-like problems to characterize their distinct problem-solving styles. We show that the models differ not only in overall performance but also in how they handle problems from various mathematical categories, in the distribution of scores they achieve, and in whether they provide complete, rigorous solutions or merely sketches. 
The Gemini models demonstrated the most mature mathematical behavior, frequently producing detailed and fully justified proofs, with few intermediate or partially correct responses. In contrast, the OpenAI's models exhibited more exploratory reasoning: they tended to identify key ideas and partial arguments but frequently failed to develop them into complete, formally correct proofs. The Claude Sonnet model showed a similar profile, characterized by coherent and readable solutions that occasionally contained conceptual inaccuracies or omitted critical steps. The DeepSeek's model, on the other hand, displayed the weakest performance, typically offering sketches or outlines without sufficient justification or mathematical depth.

Our analysis also allowed us to assess the difficulty of the Putnam-like dataset itself. Based on a comparison of the models' performance on this dataset with their results on authentic problems from the 2024 Putnam competition, we find that the difficulty level of the prepared benchmark is slightly lower than that of the actual competition.

\section{Statistics and analysis of the results}

\subsection{Total distribution of grades}
\begin{figure}[H]
\centering
\includegraphics[width=0.8\textwidth]{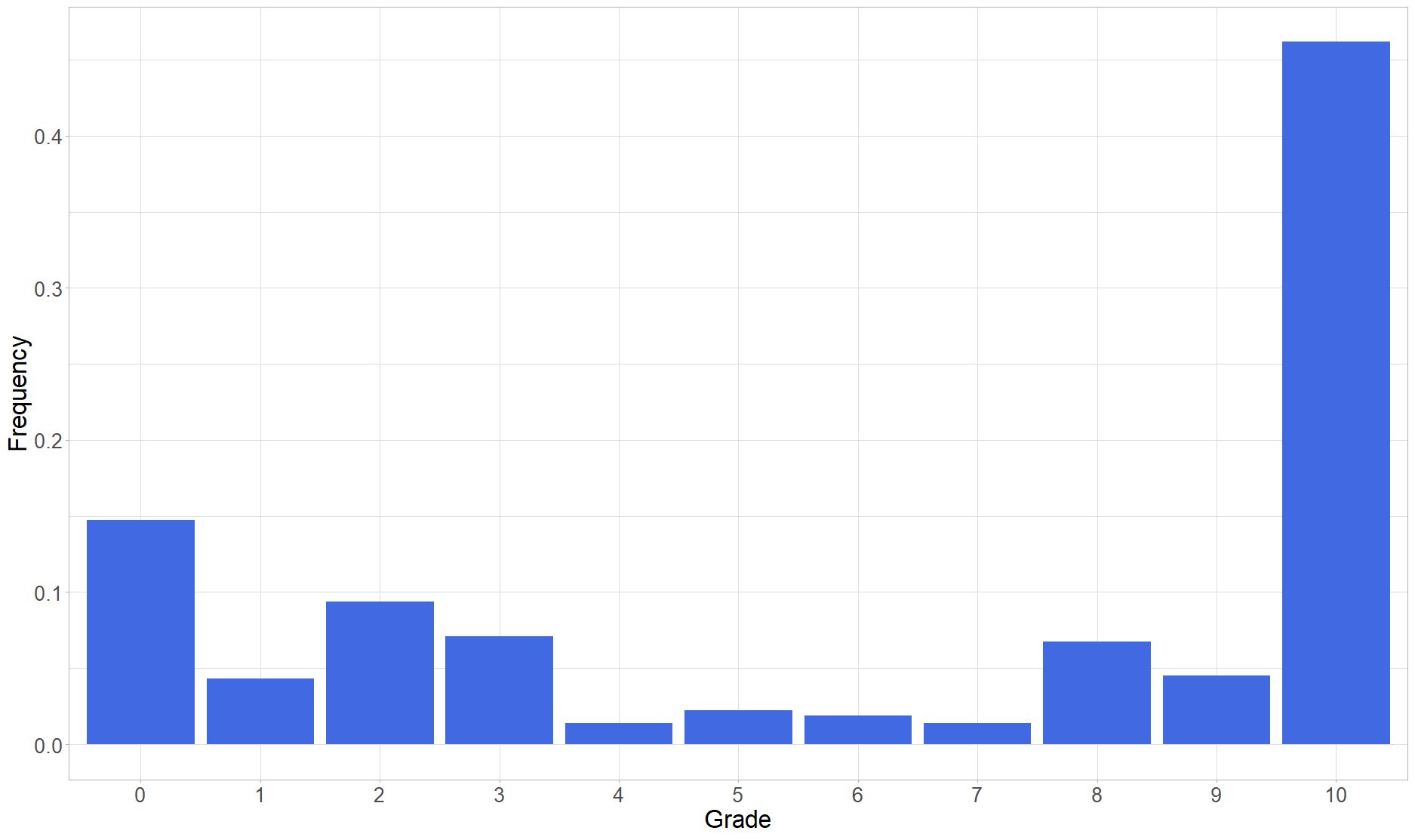}
\caption{Total distribution of grades}
\label{f1}
\end{figure}

The $576$ solutions in the dataset were graded with integers from $0$ to $10$. The frequencies of each grade are given in Table \ref{t1} and visualized in Figure \ref{f1}. 

\begin{table}[h!]
\centering
\begin{tabular}{|c||c|c|c|c|c|c|c|c|c|c|c|c|}
\hline
Grade & 0 & 1 & 2 & 3 & 4 & 5 & 6 & 7 & 8 & 9 & 10 \\ \hline
Frequency & 0.15 & 0.04 & 0.09 & 0.07 & 0.01 & 0.02 & 0.02 & 0.01 & 0.07 &0.05 & 0.46 \\ \hline
\end{tabular}
\caption{Total distribution of grades}
\label{t1}
\end{table}

Most problems from the dataset were solved correctly by the models. This result is consistent with findings from other benchmarks, which show that the actual models are very good at solving problems from competitions for students. For example, in MathArena \cite{x1}, the models achieved scores exceeding 90\% on the problems from IMC 2025. Note that the performance of the models analyzed in MathArena on IMO 2025 problems is much lower -- at most 38.1\%. This is due to the different characteristics of competitions for school and university students. Student competition tasks are mostly based on creative proficiency in academic topics that are well mastered by LLMs, while the IMO problems often require more originality than advanced methods. 
However, despite the high difficulty of the IMO 2025 problems, state-of-the-art models demonstrated strong performance within the prescribed time limits. The advanced release of Google’s Gemini Deep Think \cite{xx2}, Harmonic’s Aristotle \cite{xx3}, and an experimental model developed by OpenAI each attained gold-medal performance, successfully solving 5 out of 6 problems. In comparison, ByteDance’s Seed-Prover \cite{x11} reached the silver-medal threshold with solutions to 4 problems.

\begin{figure}[H]
\centering
\includegraphics[width=0.8\textwidth]{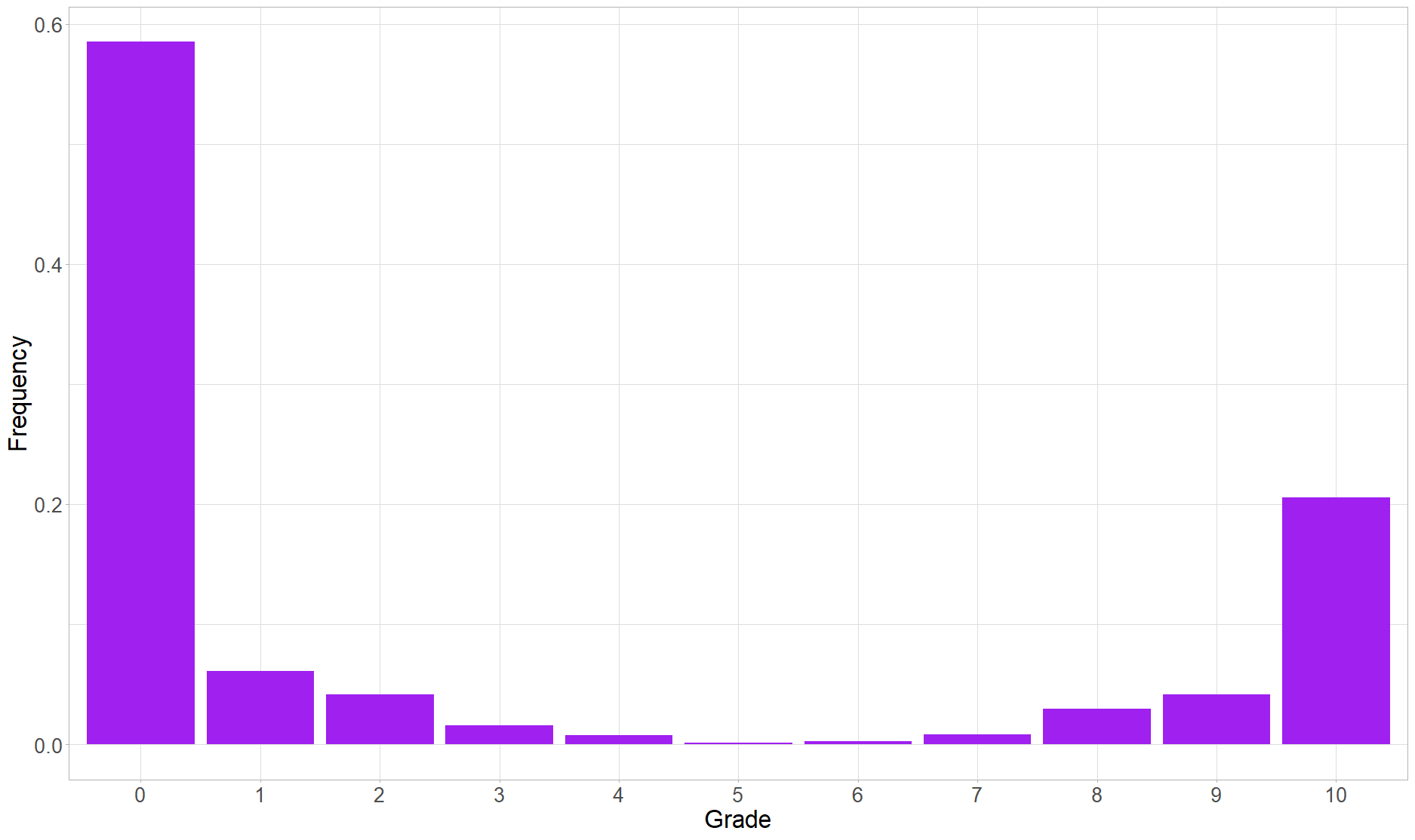}
\caption{Total distribution of grades in the Putnam competition (best contestants from the last 4 years)}
\label{f2}
\end{figure}

The Putnam competition consists of problems on different levels of difficulty, and although the easiest problems are always expected to be done by most contestants, the hardest ones are solved very rarely. The distribution of the grades of the best contestants (about 500 leading results each year) on the Putnam competition in the last four years is given in Figure \ref{f2}, see \cite{xx1}.

Note also that in the dataset the grades from 3 to 7 are more frequent than in the Putnam competition. This distribution was a consequence of using rubrics given as steps of the original solution that were needed for the automatic grading. The experts were bound by these guidelines, and for this reason, more partial points than in the actual Putnam competition were expected.

In this paper we study the grades given for the natural language solutions prepared by LLMs, so it is a significantly different class of solutions than the human ones. The only comparable dataset is the Putnam-AXIOM benchmark \cite{x2}, where the results are low and the best one was obtained by \emph{o1-preview} model with 41.94\% correct answers. However, there is a crucial difference between Putnam-AXIOM and our study. They focus on the correct boxed answer to binary verify if the problem is solved or not, while in the Putam-like dataset the entire reasoning was verified and partial grades could be awarded.

To obtain a reliable comparison of the difficulty of the tasks in the dataset, the authors evaluated the solutions of the same six models on tasks from the 2024 Putnam competition, using the same grading criteria (we do not publish the model's answers and our reviews). Note that we did not know the grading schemes used in the actual Putnam competition, so this grades cannot be directly compared with Putnam results.

The results by model are presented in Table \ref{tab-putnam-scores} together with average scores on the dataset. 

\begin{table}[h!]
\centering
\resizebox{\textwidth}{!}{
\begin{tabular}{|c||c|c|c|c|c|c|}
\hline
Model & \textit{gemini-2.5-flash-04-17} & \textit{gemini-2.5-pro-03-25}  & \textit{o3-mini-high} & \textit{o4-mini-high} & \textit{r1} & \textit{sonnet-3.7}  \\
\hline \hline
Putnam 2024&  4.3 & 7.2 &  3.6 & 3.7 & 1.1 & 3.8 \\
\hline
Putnam-like dataset &  7.6 & 8.7 &  5.6 & 6.0 & 4.5 & 6.0 \\
\hline

\end{tabular}
}
\caption{Average scores obtained by models}
\label{tab-putnam-scores}
\end{table}

Table \ref{tab-putnam-scores} can be considered as a reference point to describe the general difficulty of the Putnam-like dataset. Based on this premise, we conclude that it is rather easier than the actual Putnam competition.

\subsection{Results by problem level}

The distributions of the grades we visualize using well-known boxplots and dotplots (if the sample is small).

\begin{figure}[H]
\centering
\includegraphics[width=0.8\textwidth]{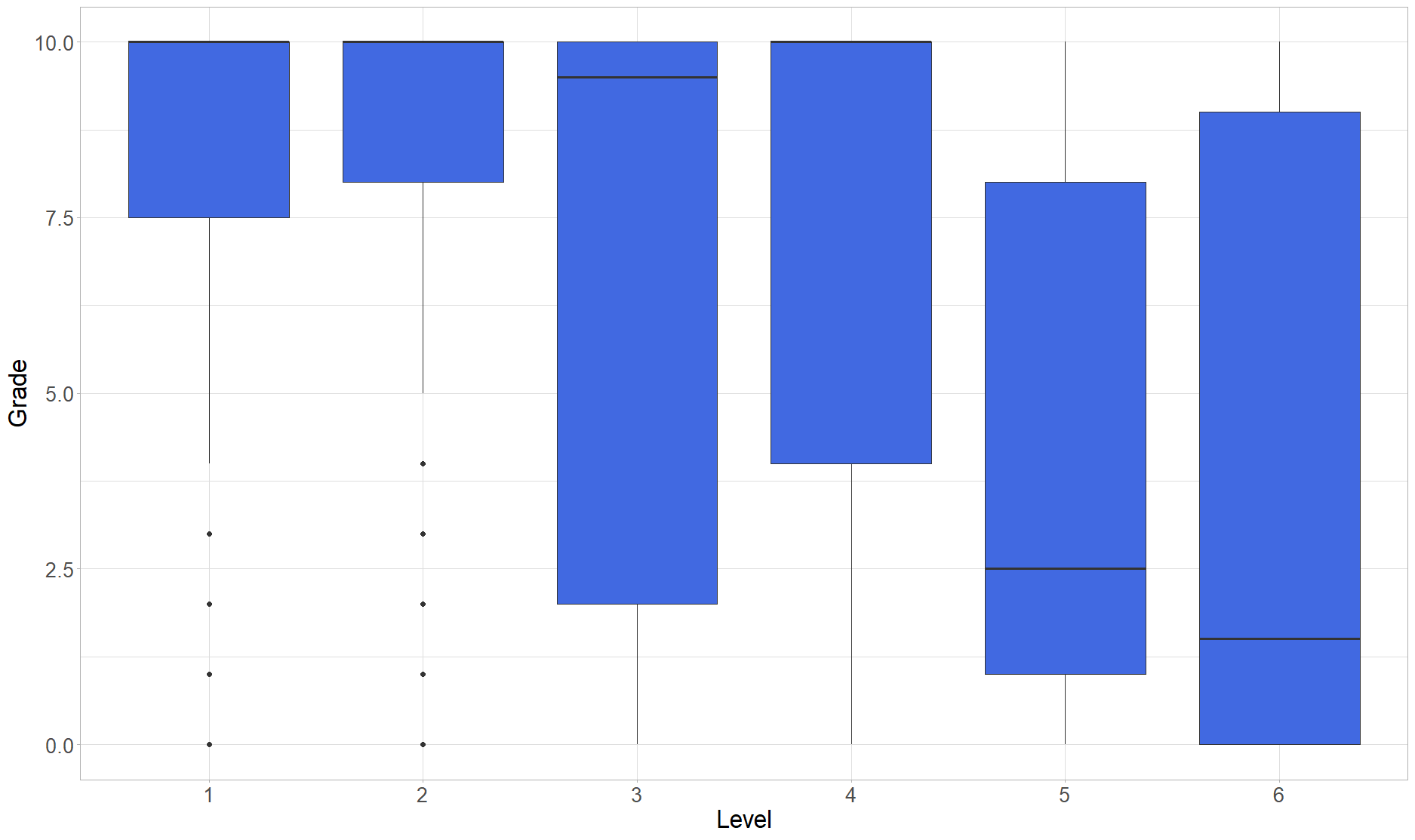}
\caption{Distribution of grades by problem level}
\label{f3}
\end{figure}

As can be seen in Figure \ref{f3}, in overview, the problems are ordered in increasing difficulty, and the problems at levels 5-6 were significantly harder for LLMs than the rest. However, level 4 problems are easier for models than level 3 problems. Welch's two sample $t$-test confirms that this difference is statistically significant ($p$-value of one-sided test is equal to 0.02). 

This anomaly may be attributed to human-perceived problem difficulty. The same thing one can observe in the results from the actual Putnam competition. In fact we should consider the problems as easy (level 1 and level 2), medium (3 and 4) and hard (5 and 6) where the order in each group is not important. This also highlights a potential discrepancy between human-perceived difficulty and the actual challenges faced by LLMs, suggesting that models may excel at problems requiring extensive calculations or knowledge recall, which humans might rate as more difficult.

We also observe that over 75\% of solutions of level 1-2 problems are correct or almost correct (7~points or more), and only a few solutions are wrong.

\subsection{Results by category}

The number of problems in each category can be found in Table \ref{tab-number-of-problems}.

\begin{table}[h!]
\centering
\begin{tabular}{|c||c|c|c|c|c|c||c|}
\hline
 & Level 1 & Level 2 & Level 3 & Level 4 & Level 5 & Level 6 & Total \\
\hline \hline
Linear algebra & 3 & 2 & 1 & 2 & 2 & 1 & 11 \\
\hline
Abstract algebra &  & 1 &  & 2 & 1 &  & 4 \\
\hline
Analysis & 8 & 9 & 11 & 7 & 9 & 13 & 57 \\
\hline
Discrete math. & 2 &   & 2 &   &   &   & 4 \\
\hline
Probability &   & 2 & 1 & 2 & 1 & 2 & 8 \\
\hline
Number theory & 1 &   & 1 & 2 & 2 &   & 6 \\
\hline
Polynomials & 2 & 2 &   & 1 & 1 &   & 6 \\
\hline
\end{tabular}
\caption{The number of problems in each category}
\label{tab-number-of-problems}
\end{table}

\begin{figure}[H]
\centering
\includegraphics[width=0.8\textwidth]{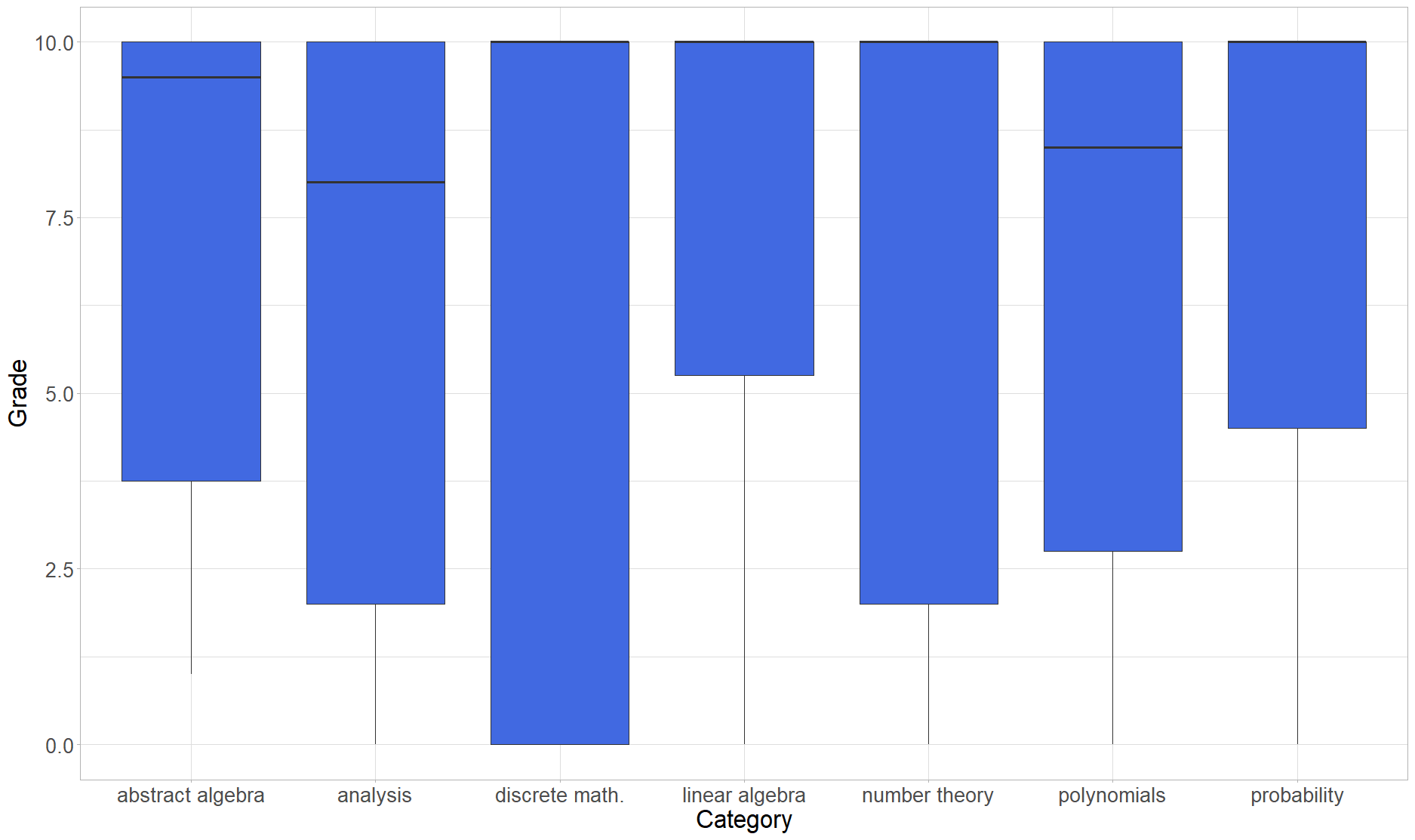}
\caption{Distribution of grades by category of the problem}
\label{f4}
\end{figure}

As we see in Figure \ref{f4}, the median is lower than 10 in three categories: abstract algebra, analysis, and polynomials. The explanation of this fact requires a detailed study. We suspect that, in the case of mathematical analysis, some of the models did not get the maximum score because of the use of numerical computations and unjustified approximations, although the final answer was correct. The plot also shows that problems in linear algebra were easier for models than the rest of the problems.

\subsection{Statistics of results by model}

\begin{figure}[H]
\centering
\includegraphics[width=0.8\textwidth]{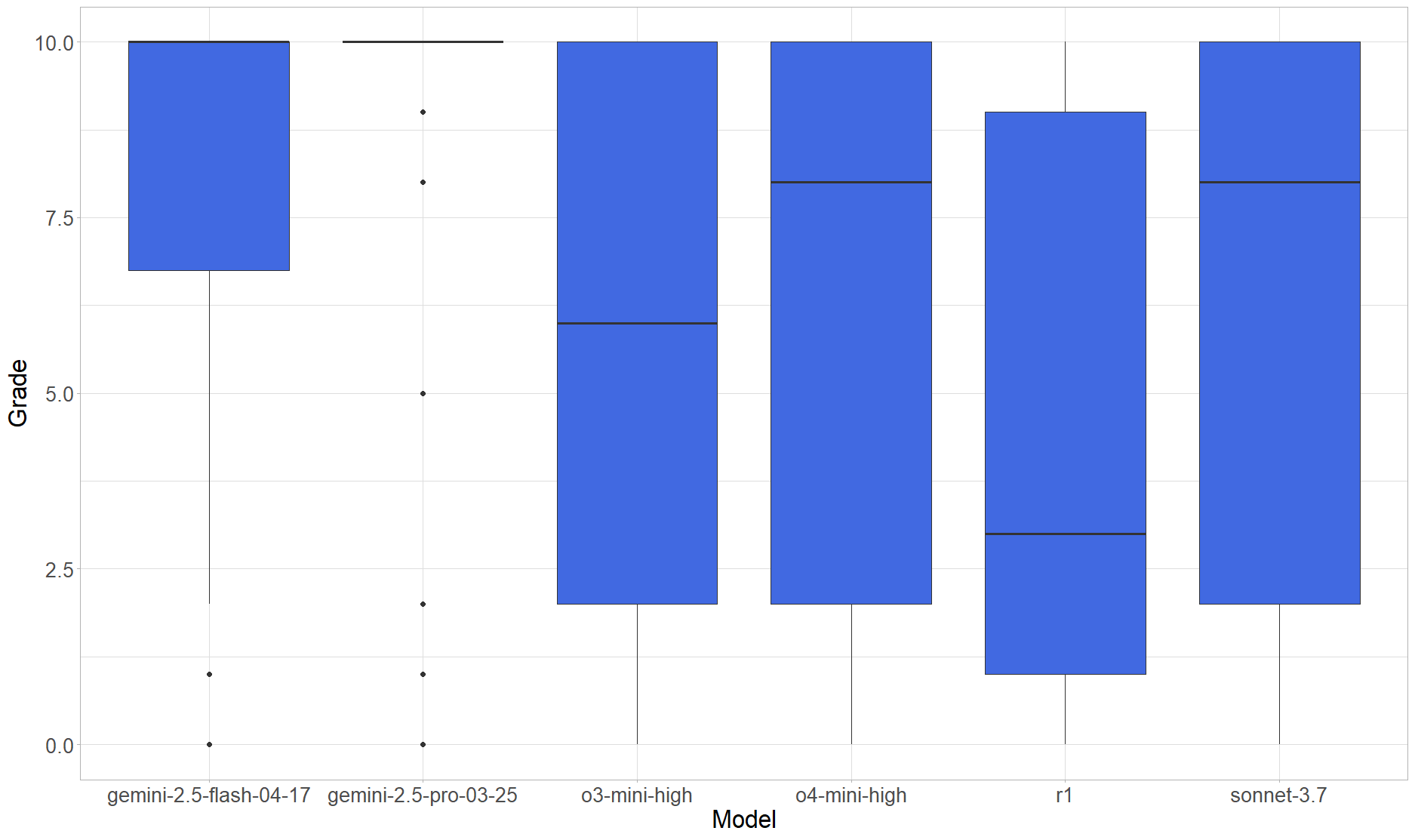}
\caption{Distribution of grades by model}
\label{fig-by-model}
\end{figure}

Significantly, the best grades were scored by \textit{gemini-2.5-pro-03-25}, and the efficiency of its faster counterpart \textit{gemini-2.5-flash-04-17} is also very good, see Figure \ref{fig-by-model}. On the other hand, the \textit{r1} model got the lowest grades. A brief analysis of the dataset reveals that the \emph{r1} model frequently provides only a sketch of a solution, without providing any details or justifications. Such a behavior is possible due to the lack of any additional prompt explaining that the solutions should contain all the details and justifications. 

Observe that distributions of grades scored by \emph{o4-mini-high} and \emph{sonnet-3.7} are very similar. It was also the case in Table \ref{tab-putnam-scores}. Kolmogorov-Smirnov test of the equality of distributions confirms this observation with $p$-value $0.998$.

Below we look closely at the grades scored by each model, compared by category and the difficulty level of the problem. Since the number of problems in most of the categories is small, we do not use boxplots for the comparison by category.

\begin{figure}[H]
\centering
\begin{subfigure}{0.45\textwidth}
    \centering
    \includegraphics[width=\textwidth]{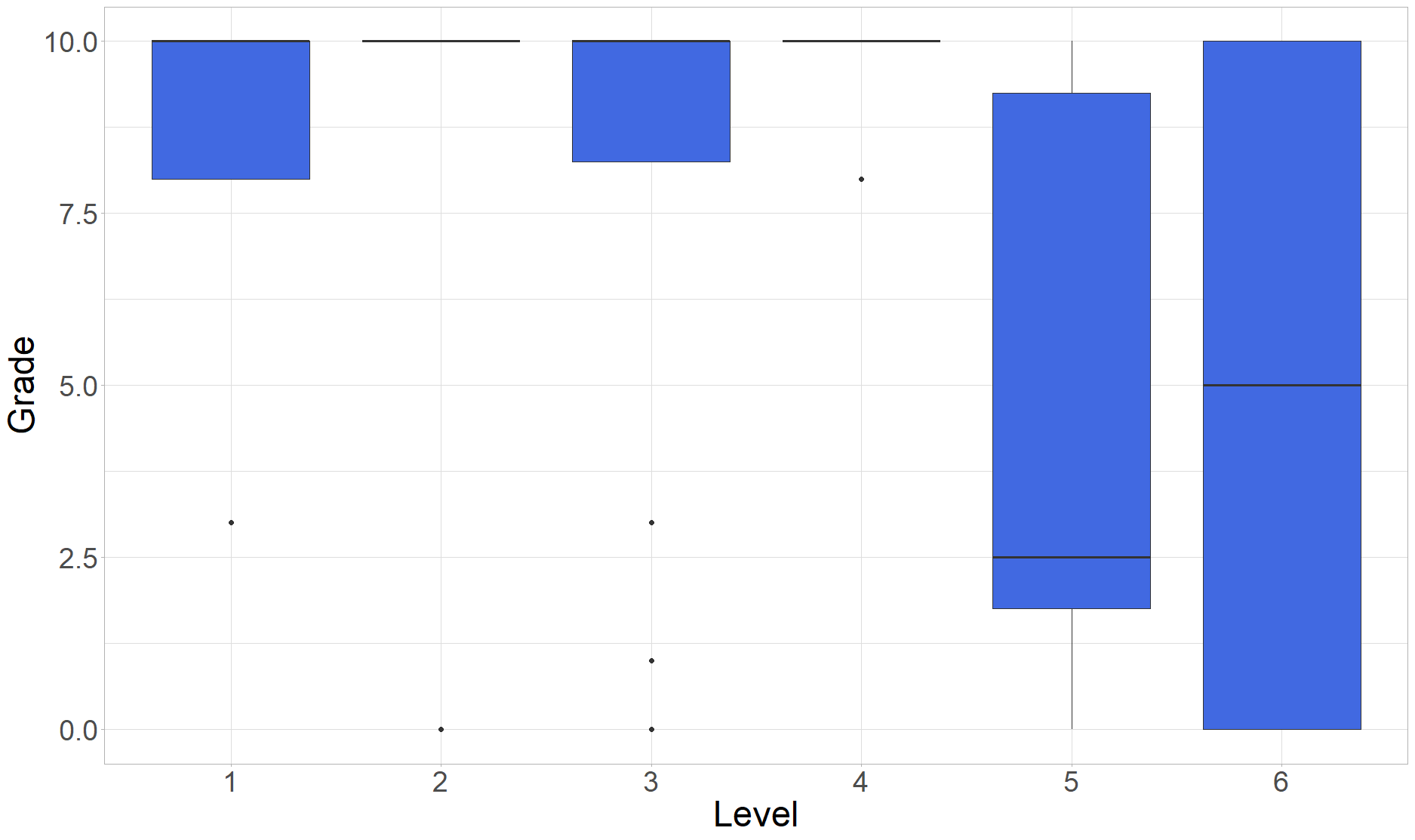}
    \caption{By difficulty level}
\end{subfigure}
\hfill
\begin{subfigure}{0.45\textwidth}
    \centering
    \includegraphics[width=\textwidth]{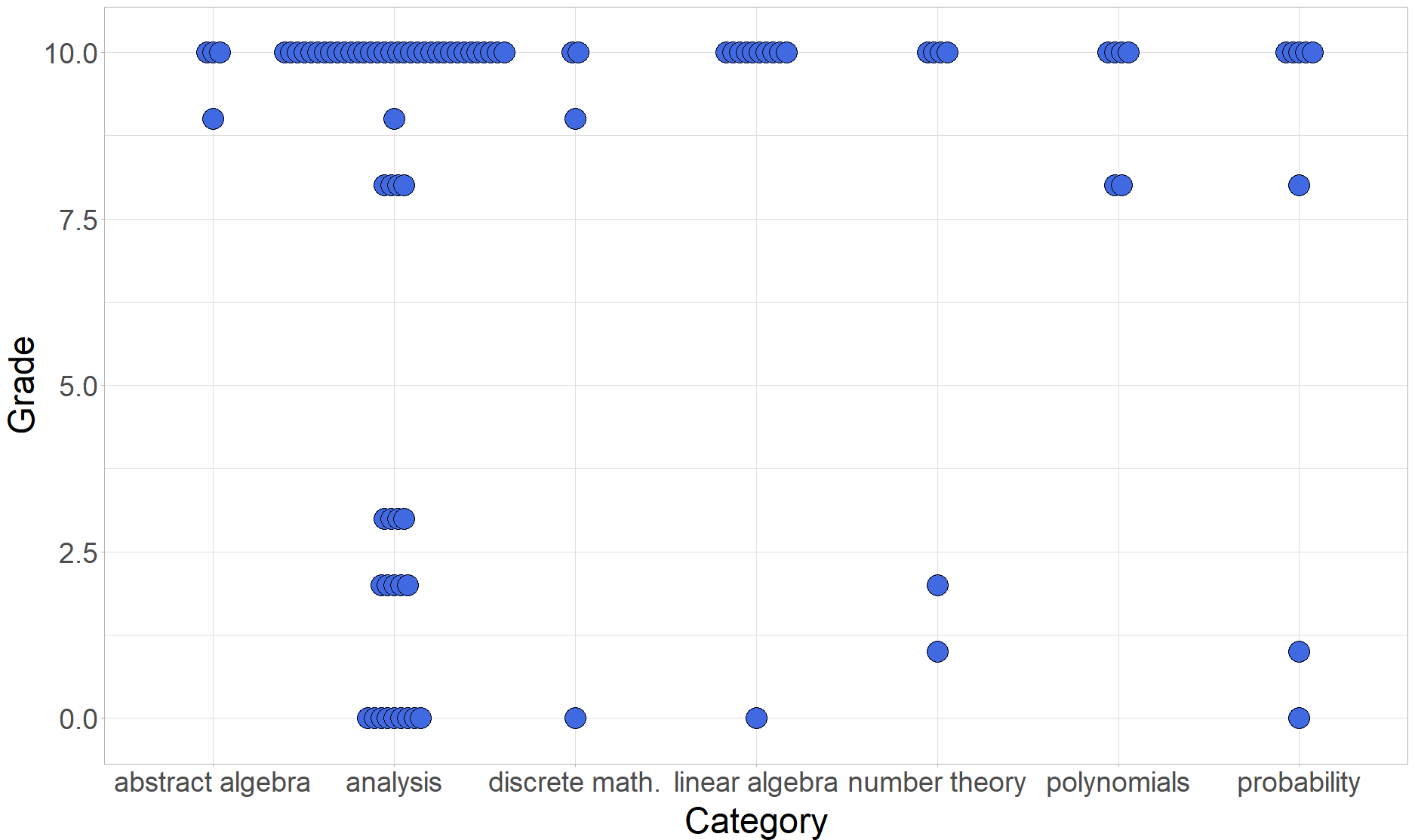}
    \caption{By category}
\end{subfigure}
\caption{Distribution of grades of \textit{gemini-2.5-flash-04-17}}
\label{fig-gemini-flash}
\end{figure}

The results of \emph{gemini-2.5-flash-04-17} are very good. However, many solutions are unnecessarily wordy, and correct arguments were mixed with plenty of dead ends in the reasoning.

\begin{figure}[H]
\centering
\begin{subfigure}{0.45\textwidth}
    \centering
    \includegraphics[width=\textwidth]{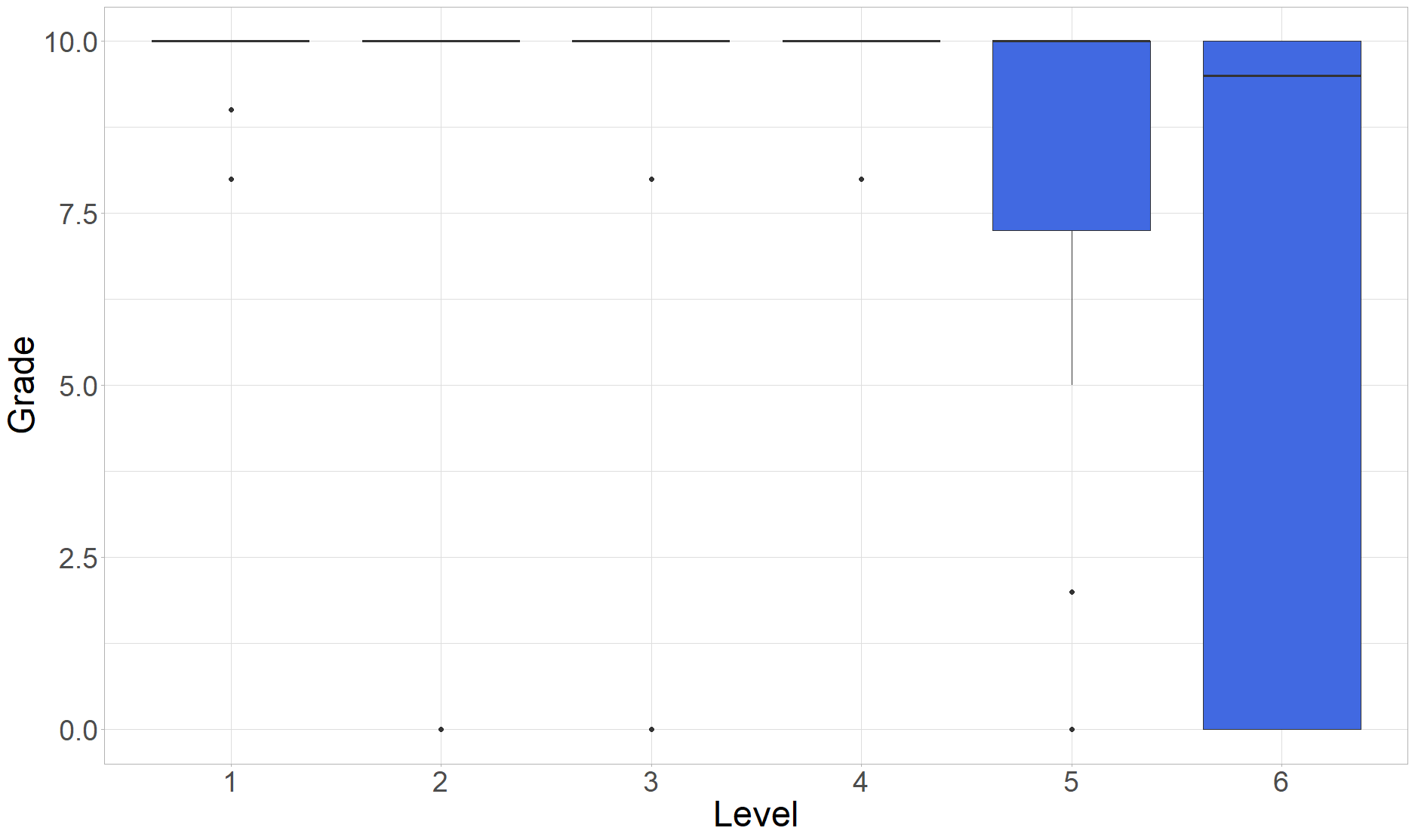}
    \caption{By difficulty level}
\end{subfigure}
\hfill
\begin{subfigure}{0.45\textwidth}
    \centering
    \includegraphics[width=\textwidth]{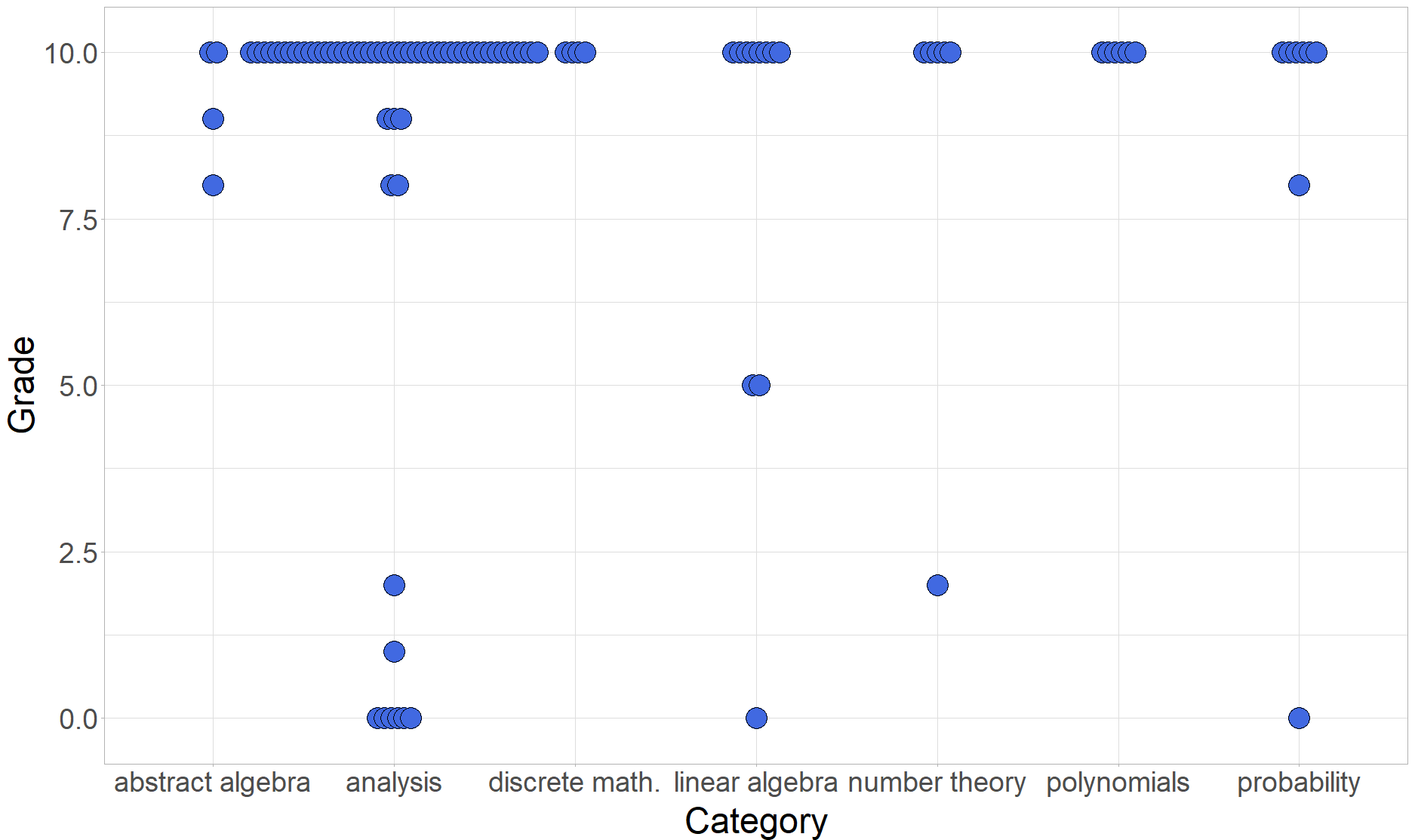}
    \caption{By category}
\end{subfigure}
\caption{Distribution of grades of \textit{gemini-2.5-pro-03-25}}
\label{fig-gemini-pro}
\end{figure}

 Observe that \textit{gemini-2.5-pro-03-25} is the only model whose median of grades in problems at levels 5-6 is close to $10.0$ (exactly $10.0$ at level 5 and $9.5$ at level 6). Still, there are hard problems in which it scores a low number of points. Clearly, most of the unsolved problems belong to the category of mathematical analysis (which might be the property of the dataset, as over half of the problems belong to this category). 
 
Note that most solutions of this model were completely correct or completely incorrect; the number of partially correct solutions is very small. This means that if \textit{gemini-2.5-pro-03-25} is able to provide the correct answer, the reasoning is rigorous and complete.

\begin{figure}[H]
\centering
\begin{subfigure}{0.45\textwidth}
    \centering
    \includegraphics[width=\textwidth]{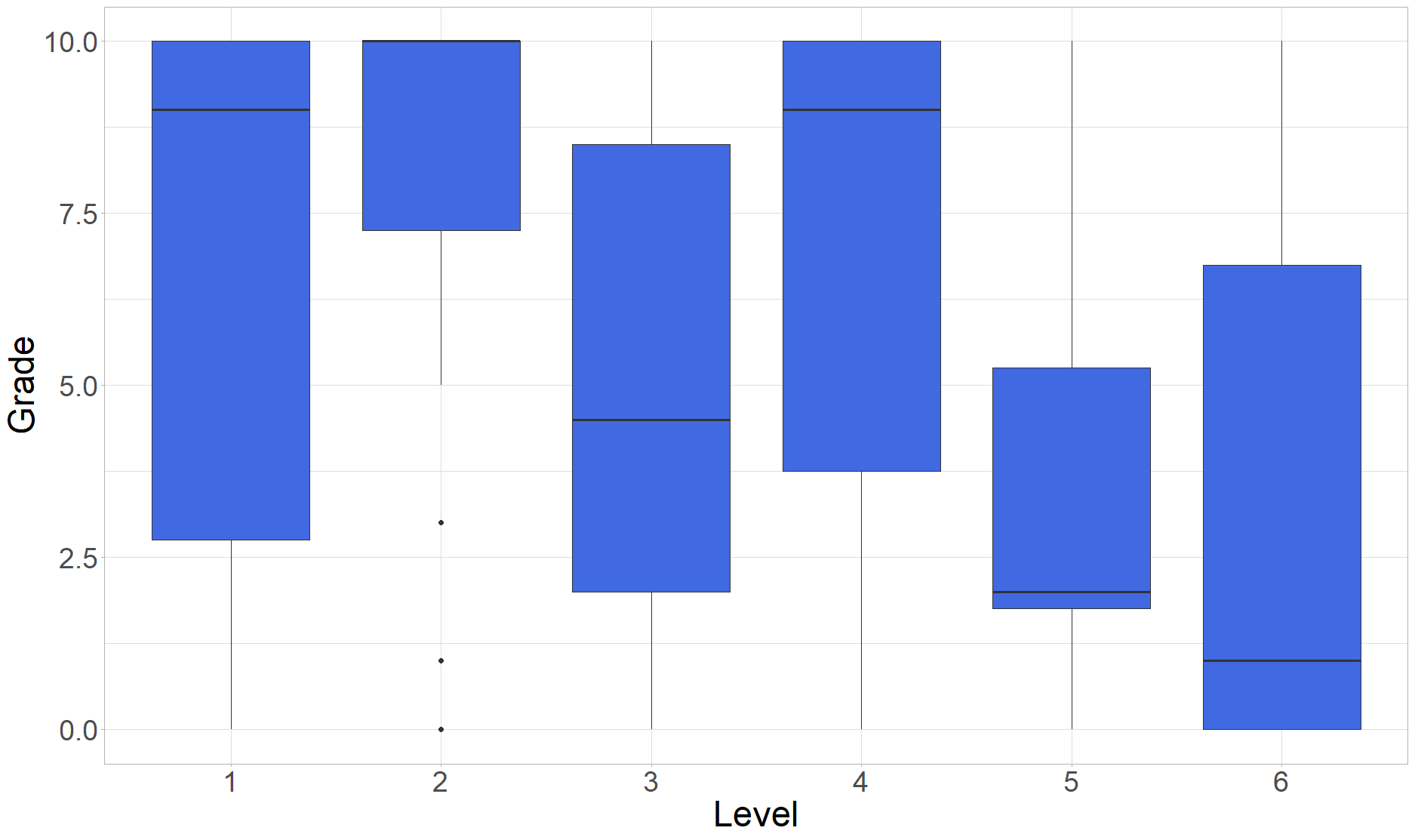}
    \caption{By difficulty level}
\end{subfigure}
\hfill
\begin{subfigure}{0.45\textwidth}
    \centering
    \includegraphics[width=\textwidth]{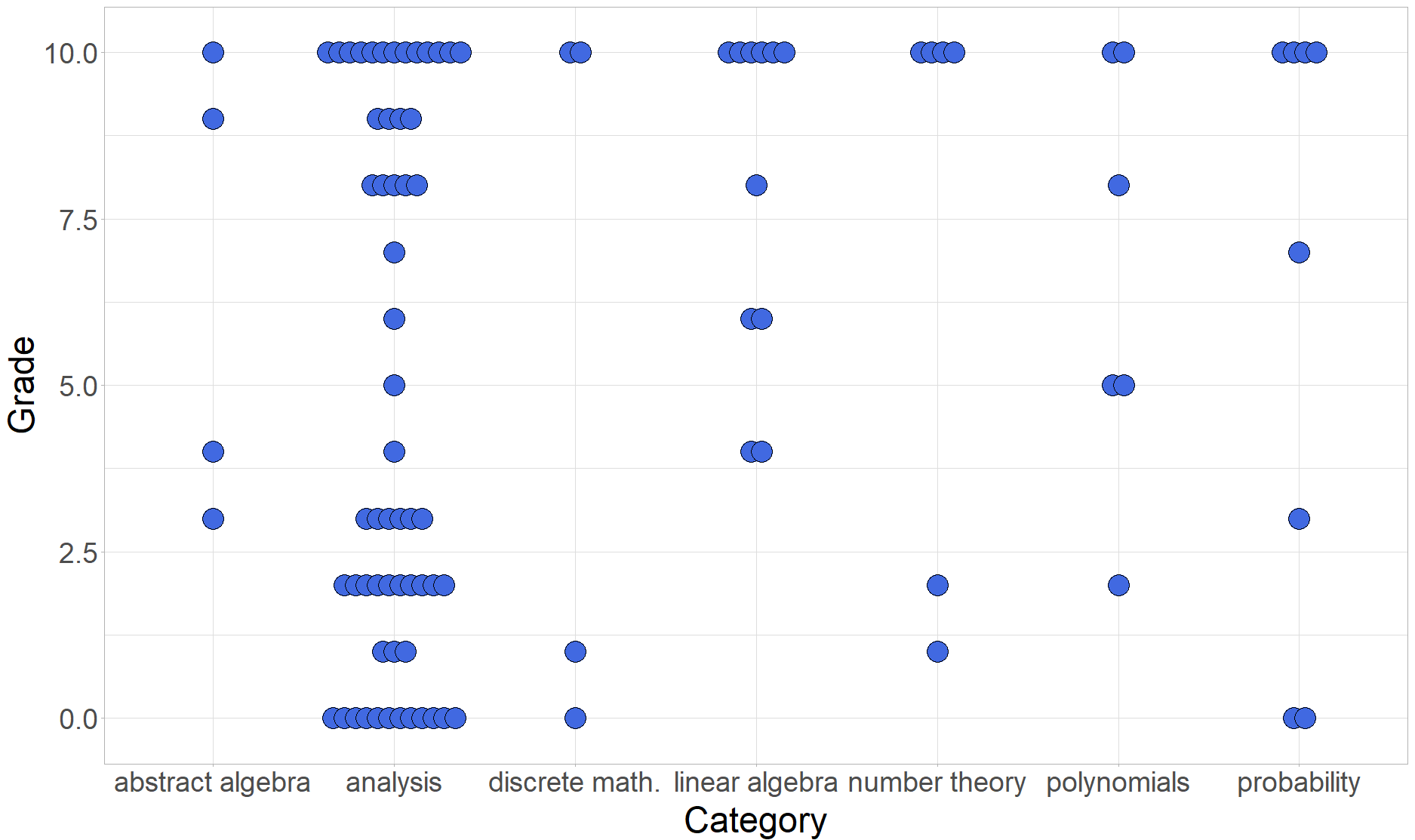}
    \caption{By category}
\end{subfigure}
\caption{Distribution of grades of \textit{o3-mini-high}}
\label{fig-o3}
\end{figure}

Observe that the \emph{o3-mini-high} model got a lot of partial points, i.e., grades from 2 to 8. This is the case of 41\% of solutions. This means that this model proposes correct ideas but often cannot extend them to a full, rigorous solution.

\begin{figure}[H]
\centering
\begin{subfigure}{0.45\textwidth}
    \centering
    \includegraphics[width=\textwidth]{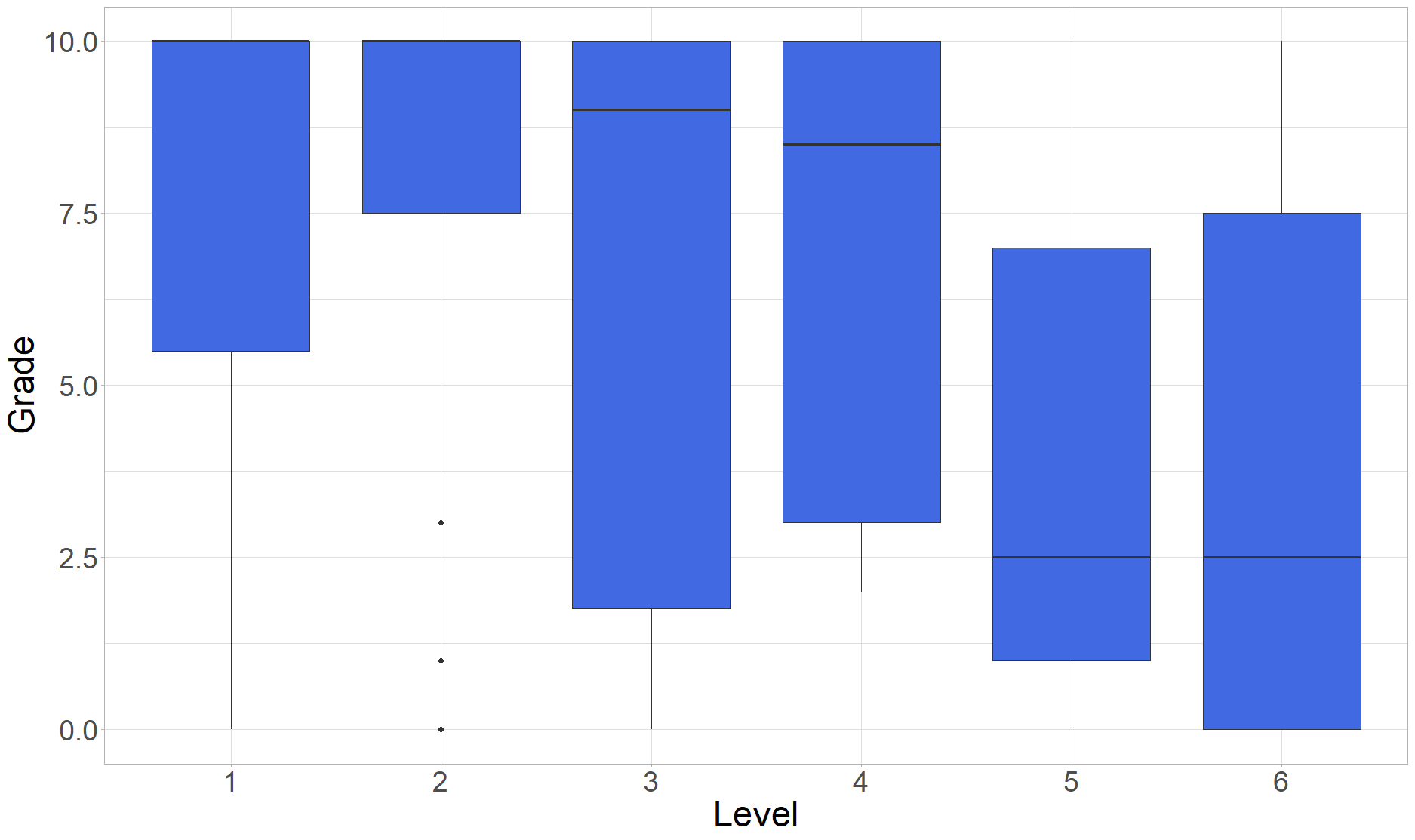}
    \caption{By difficulty level}
\end{subfigure}
\hfill
\begin{subfigure}{0.45\textwidth}
    \centering
    \includegraphics[width=\textwidth]{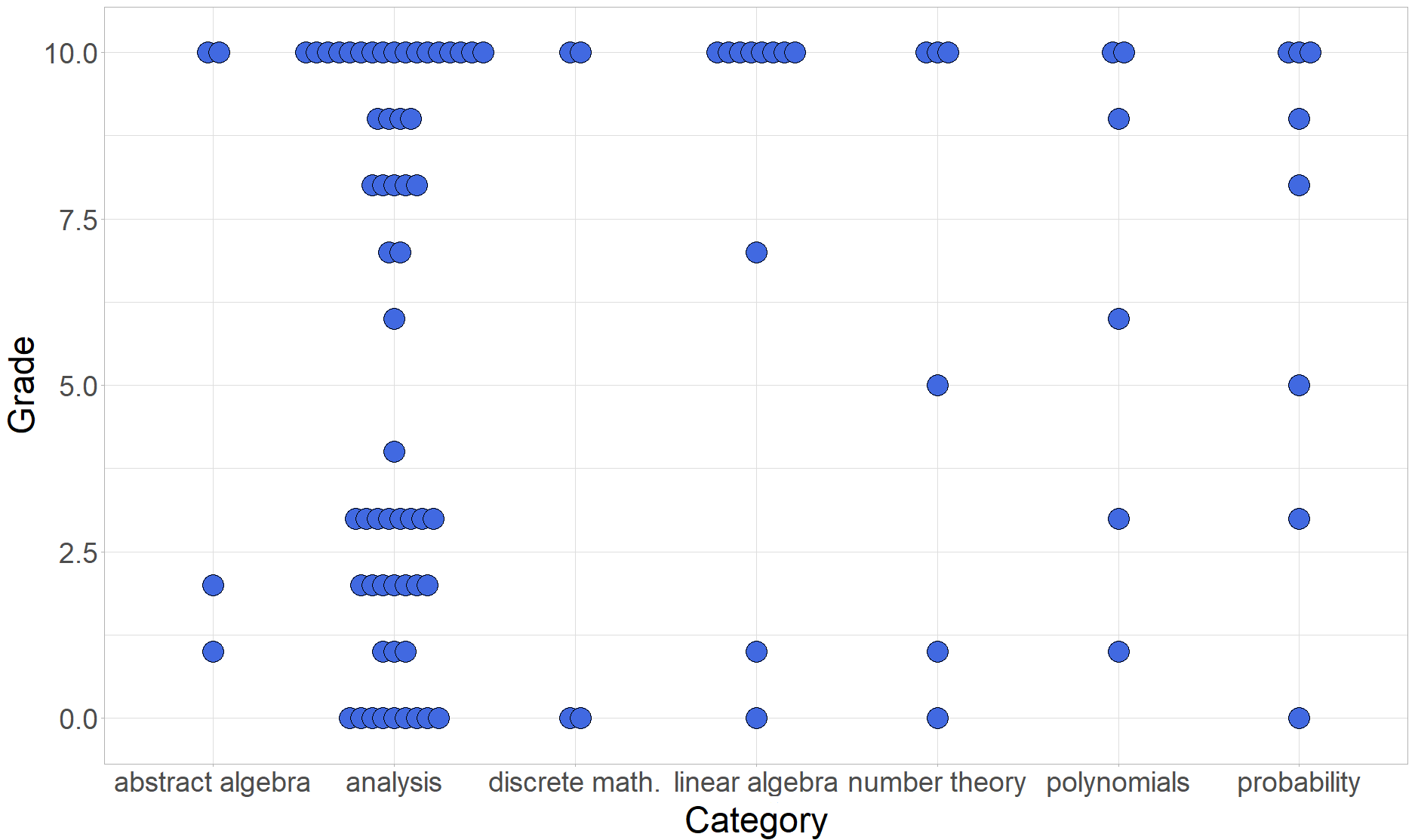}
    \caption{By category}
\end{subfigure}
\caption{Distribution of grades of \textit{o4-mini-high}}
\label{fig-o4}
\end{figure}

The boxplots of \emph{o3-mini-high} and \emph{o4-mini-high} are similar, and the $t$-test doesn't distinguish the average of these two samples. However, a closer look shows a significant difference between these models. Observe that \emph{o4-mini-high} has a greater median (cf. Figure \ref{fig-by-model}) and a smaller number of partially correct solutions. It means that \emph{o4-mini-high} shows a greater ability to provide full mathematical solutions than the previous generation.

\begin{figure}[H]
\centering
\begin{subfigure}{0.45\textwidth}
    \centering
    \includegraphics[width=\textwidth]{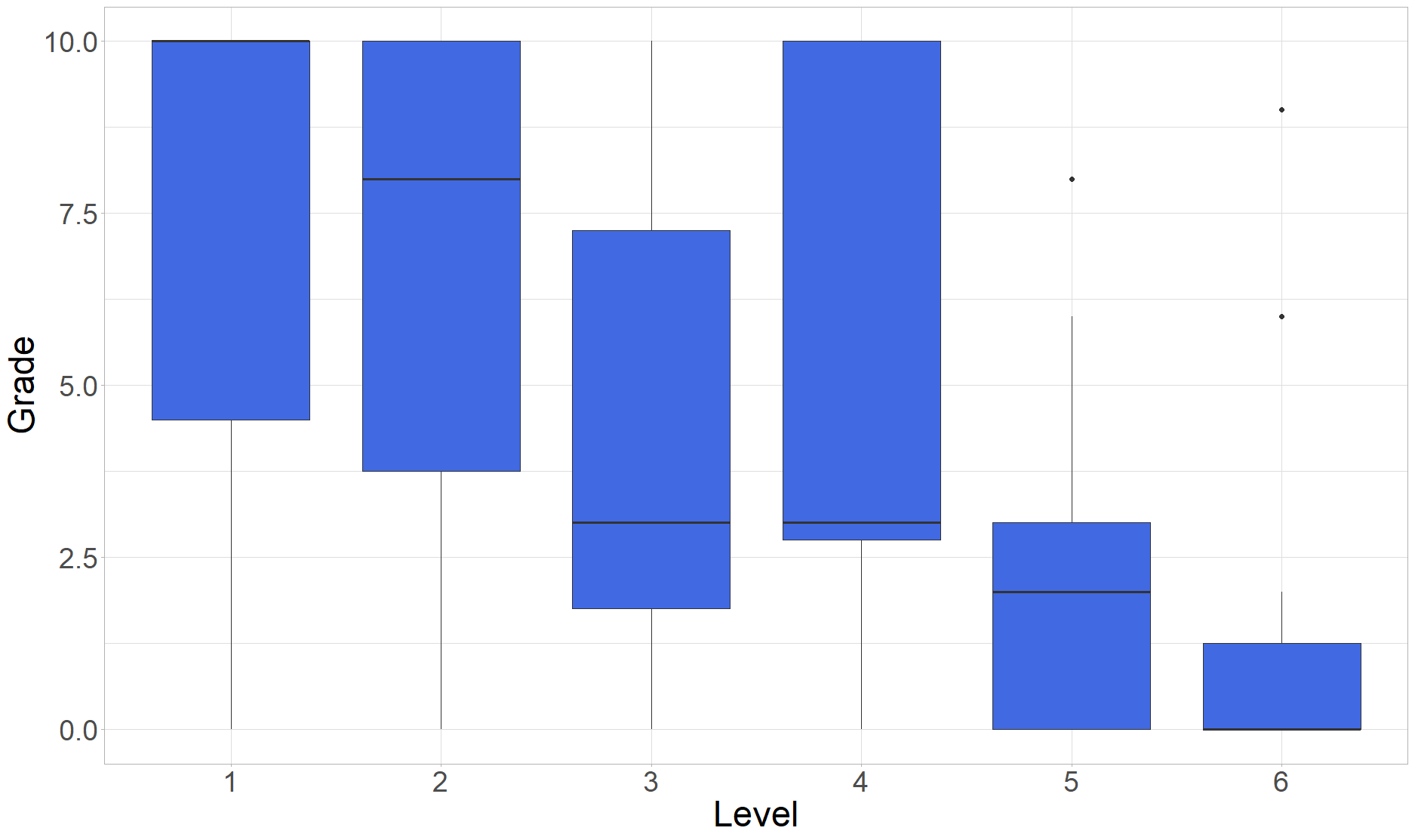}
    \caption{By difficulty level}
\end{subfigure}
\hfill
\begin{subfigure}{0.45\textwidth}
    \centering
    \includegraphics[width=\textwidth]{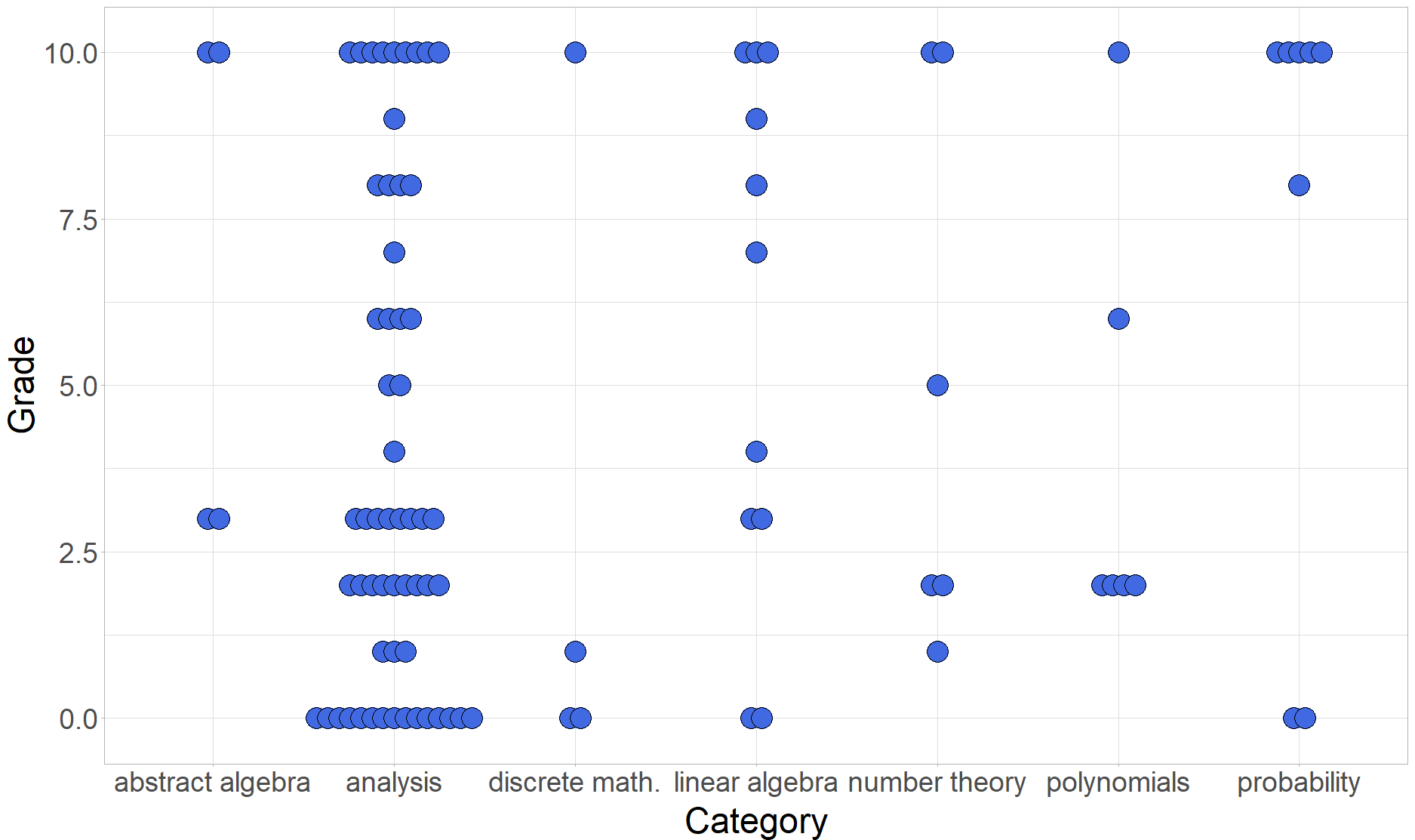}
    \caption{By category}
\end{subfigure}
\caption{Distribution of grades of \textit{r1}}
\label{fig-r1}
\end{figure}

For the \emph{r1} model, only solutions to problems at level 1 were graded with a median $10.0$, which shows that the model, in general, doesn't score high grades in most of the problems; in fact, in the problems with level 3 and higher, the median is significantly smaller than $5.0$. A closer look at the solutions provided by the model shows that it produces only sketches of solutions, without detailed calculations or rigorous arguments.

\begin{figure}[H]
\centering
\begin{subfigure}{0.45\textwidth}
    \centering
    \includegraphics[width=\textwidth]{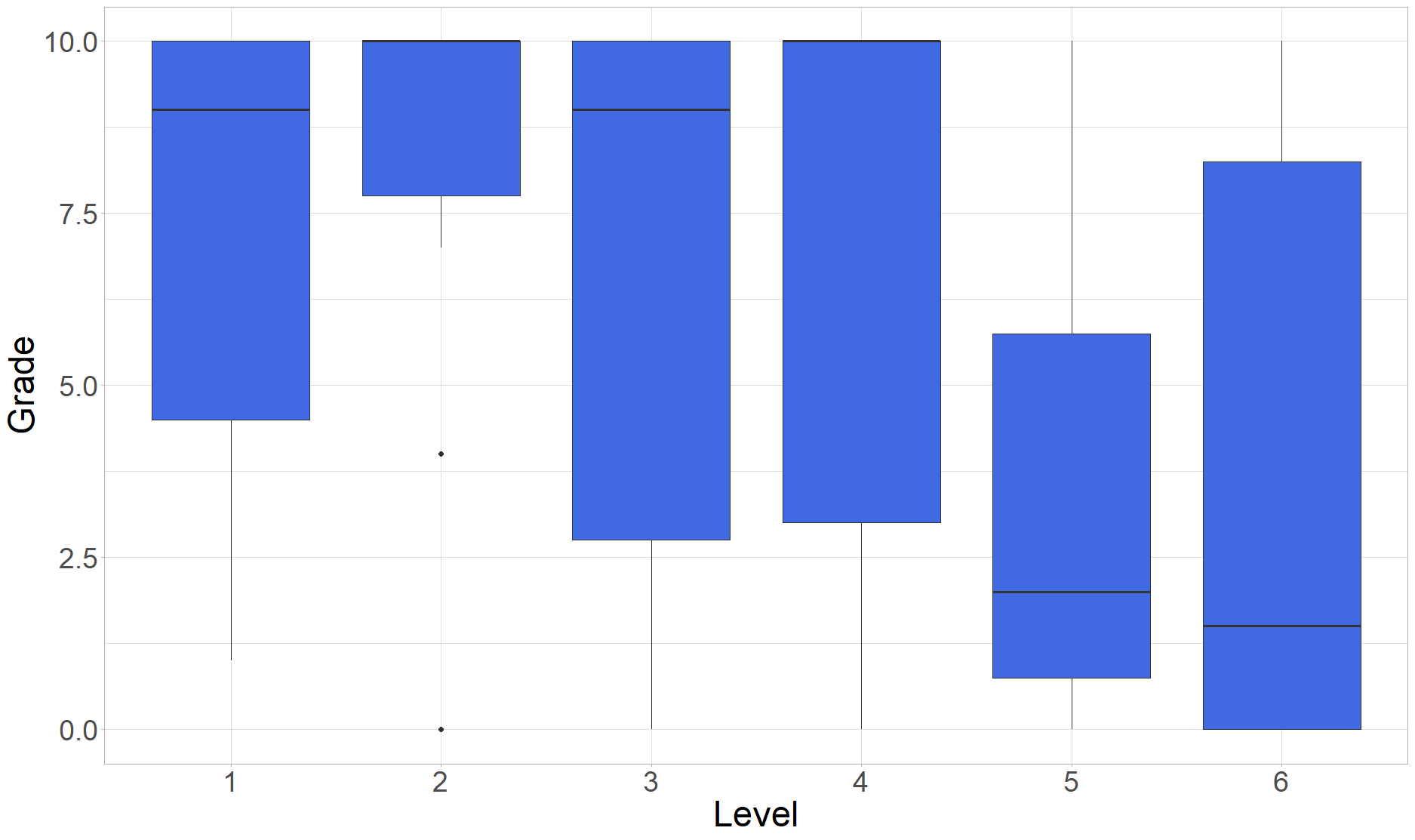}
    \caption{By difficulty level}
\end{subfigure}
\hfill
\begin{subfigure}{0.45\textwidth}
    \centering
    \includegraphics[width=\textwidth]{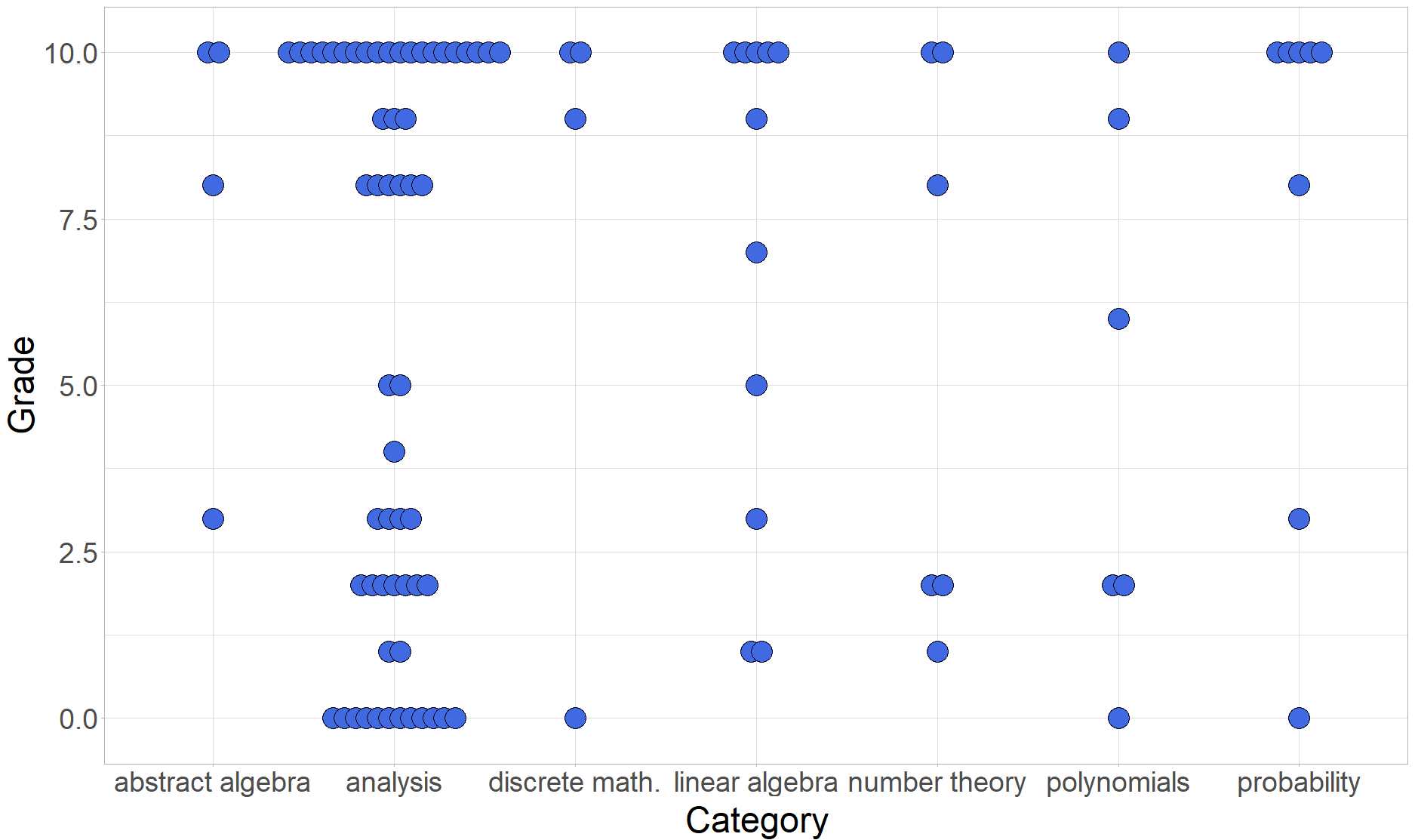}
    \caption{By category}
\end{subfigure}
\caption{Distribution of grades of \textit{sonnet-3.7}}
\label{fig-sonnet}
\end{figure}

The results of \emph{sonnet-3.7} are similar to those of \emph{o3-mini-high} and \emph{o4-mini-high}. It is worth remarking, that solutions of this model are well-edited and easy to read.

To conclude, we provide a summary of the average grades of each model by category.
\begin{table}[h!]
\centering
\resizebox{\textwidth}{!}{
\begin{tabular}{|c||c|c|c|c|c|c|c|c|}
\hline
Model & Average score & Abstract algebra  & Linear algebra  & Analysis  & Discrete math.  & Probability  & Number theory  & Polynomials  \\
\hline \hline
\textit{gemini-2.5-flash-04-17} & $7.59$ & $9.75$ & $9.09$ & $7.07$ & $7.25$ & $7.38$ & $7.17$ & $9.33$ \\
\hline
\textit{gemini-2.5-pro-03-25} & $8.68$ & $8.18$ & $9.25$ & $8.53$ & $10.0$ & $8.50$ & $8.67$ & $10.0$  \\
\hline
\textit{o3-mini-high} & $5.59$ & $8.00$ & $6.50$ & $4.72$ & $5.25$ & $6.25$ & $7.17$ & $6.67$  \\
\hline
\textit{o4-mini-high} & $5.95$ & $8.00$ & $5.75$ & $5.46$ & $5.00$ & $6.88$ & $6.00$ & $6.50$  \\
\hline
\textit{r1} & $4.52$ & $5.82$ & $6.50$ & $3.88$ & $2.75$ & $7.25$ & $5.00$ & $4.00$  \\
\hline
\textit{sonnet-3.7} & $6.00$ & $6.91$ & $7.75$ & $5.56$ & $7.25$ & $7.63$ & $5.50$ & $4.83$  \\
\hline
\end{tabular}
}
\caption{Average scores}
\label{tab-average-scores}
\end{table}

\section{Conclusions}
Using the Putnam-like dataset, we effectively simulated a mathematics competition modeled on the William Lowell Putnam Mathematical Competition, with large language models (LLMs) as participants. The difficulty level of the problems was slightly lower than in last year’s edition of the contest.

Our analysis shows that the models behave in ways comparable to human contestants: they often demonstrate good intuition, but also make classical mistakes and present flawed solutions. The strongest performances were achieved by \emph{gemini-2.5-pro-03-25} and \emph{gemini-2.5-flash-04-17}, which produced the most detailed solutions. By contrast, the \emph{r1} model exhibited the highest error rate and the weakest overall results. Nevertheless, all six evaluated models performed remarkably well when compared to actual human outcomes in the competition.

Future work will focus on a deeper analysis of the details of the models' solutions and the errors committed. Another important aspect is the correlation between human and automatic grading, which could lead to the development of more reliable evaluation systems, as well as exploring the impact of advanced prompting techniques on the models' ability to generate rigorous mathematical proof.

\appendix
\section{Case study}
In what follows we describe specific behavior of the models in four selected problems to show how they differ in their approach, as it was described in overview in Conclusions. The full answers given by models are available in the GitHub repository.

\begin{center}
\begin{tcolorbox}[colback=white,colframe=black,boxrule=0.4pt,
arc=4mm,enhanced,width=0.9\textwidth,sharp corners=south]
\textbf{\textit{Set 1, Problem A1}}

\begin{tcolorbox}[colback=lightgray,colframe=black,arc=1mm,boxrule=0.2pt]
Let $\mathbb{P}$ denote the set of all primes. Define
 $$
 \mathcal{D} := \{ pq \ : \ p,q \in \mathbb{P} \}.
 $$
 Find the maximal length of a sequence of censecutive integers in $\mathcal{D}$.

\end{tcolorbox}
\url{https://github.com/google-deepmind/eval_hub/tree/master/eval_hub/putnam_like/putnam_like/Set_1/A1}

\end{tcolorbox}
\end{center}

This easy problem was correctly solved by all models. However, the solutions produced by the different models exhibit noticeable stylistic differences, both in length and in rhetorical structure. 
The longest responses are generated by \pro and \flash, each exceeding 700 words, and are characterized by extensive exposition, and a didactic, tutorial-like style. These solutions tend to guide the reader carefully through intermediate observations and often pause to explain why a particular step is justified. In contrast, the solutions produced by \othree, \ofour, and \rone are significantly shorter (approximately 200--240 words) and adopt a more concise, focusing primarily on the core argument without extended commentary.

\begin{center}
\begin{tcolorbox}[colback=white,colframe=black,boxrule=0.4pt,
arc=4mm,enhanced,width=0.9\textwidth,sharp corners=south]
\textbf{\textit{Set 1, Problem B4}}

\begin{tcolorbox}[colback=lightgray,colframe=black,arc=1mm,boxrule=0.2pt]

Find the limit
 $$
 \lim_{n\to\infty} \int_0^1 \frac{(1-\ln(1-x))^n}{n!} \, dx.
 $$

\end{tcolorbox}
\url{https://github.com/google-deepmind/eval_hub/tree/master/eval_hub/putnam_like/putnam_like/Set_1/B4}

\end{tcolorbox}
\end{center}

All evaluated models ultimately arrive at the correct result—the mathematical constant $e$. However, they demonstrate divergent methodologies, both in their purely mathematical rigor and their respective approaches to presentation. The majority of the models, including \othree, \ofour, \pro, \flash, and \sonnet, utilize the substitution $u = 1 - \ln(1-x)$, which transforms the integral into a form involving the gamma function, but on an incomplete interval $[1,\infty)$ instead of the standard $[0,\infty)$. Consequently, the primary analytical challenge in these solutions was the appropriate estimation of the integral over the interval $[0,1]$.

\pro, \flash, and \sonnet  provided precise calculations at this stage, although the Gemini models (\pro and \flash) did so in an overly verbose and excessively detailed manner.

After reducing the problem to an integral over the interval $[0,1]$, the \othree model employed an intuitive but entirely informal transition, stating: \textit{``Notice that for large \( n \) the integrand \( u^n e^{-u} \) is concentrated near \( u \approx n \), and the contribution from the interval \( 0 \le u \le 1 \) is negligible.''} This lack of formal rigor resulted in a significant loss of points for this solution.

The \ofour model utilized a formula for the incomplete gamma function. While this identity is documented (e.g., on Wikipedia), the evaluator deemed it not to be common knowledge; thus, the solver should have cited the fact or provided a derivation. It should be noted that this formula can be derived directly from the thesis of the problem itself. Interestingly, other models didn't identify or utilize this readily available formula.

The \rone model adopts a slightly different substitution in the integral, $u = -\ln(1-x)$, which leads directly to an integrand of the form $(1+u)^n e^{-u}$. This choice by \rone proves to be highly efficient, as it allows for a straightforward application of the binomial theorem. However, \rone omitted the intermediate calculations and proceeded directly to the conclusion, which precluded the solution from being graded as fully correct.

Finally, it is worth emphasizing that this task further highlights the distinct stylistic divergence of the Gemini models compared to their counterparts. \pro and \flash are highly narrative, meticulously explaining every step and providing explicit logical transitions. However, this approach may be perceived as overly prolix and difficult to digest; by focusing exhaustively on granular details, these models often fail to provide a clear conceptual overview of the solution's underlying strategy. From the perspective of a reader seeking to master the methodology of such problems, the solution provided by \sonnet would be the optimal choice, as it strikes the best balance between pedagogical clarity and structural transparency.

\begin{center}
\begin{tcolorbox}[colback=white,colframe=black,boxrule=0.4pt,
arc=4mm,enhanced,width=0.9\textwidth,sharp corners=south]
\textbf{\textit{Set 3, Problem A6}}

\begin{tcolorbox}[colback=lightgray,colframe=black,arc=1mm,boxrule=0.2pt]

Let $J_n$ denote the $n\times n$ matrix with ones in each odd column and zeros in each even column e.g. $J_3=\begin{pmatrix}
 1&0& 1 \\
 1&0& 1 \\
 1&0& 1
 \end{pmatrix}$. For which even integers $n$ does there exist an $n\times n$ matrix $A$ whose entries are all in $\{0,1\}$, such that $A^2=J_n$?

\end{tcolorbox}
\url{https://github.com/google-deepmind/eval_hub/tree/master/eval_hub/putnam_like/putnam_like/Set_3/A6}

\end{tcolorbox}
\end{center}

Problems of this nature typically necessitate a two-fold approach: first, identifying potential candidates via a necessary condition (in this case $n=2k^2$), and second, providing an explicit construction for such a matrix. None of the evaluated models successfully completed this task in its entirety. While \sonnet derived the necessary condition, its reasoning was punctuated by logical gaps and errors; furthermore, the model made no attempt to construct a matrix of the required dimensions. The \rone model quickly identified $n=2$ as a valid solution but subsequently dismissed cases where $n\geq 4$ based on laconic and erroneous arguments. Similarly, \ofour identified the trivial case $n=2$ but excluded all other possibilities due to elementary algebraic errors.

The \othree model demonstrated the necessary condition, albeit with minor analytical lapses. Although it recognized the requirement for an explicit construction, it declined to provide one, concluding with the remark: ``One may also show (by an explicit but somewhat involved combinatorial construction, see for example constructions using incidence matrices of certain designs)...'' This instance underscores a perceptible reluctance on the part of the model to engage with highly complex, constructive derivations.

The Gemini models again provided extensively detailed responses. \pro derived the necessary condition through a sophisticated yet convoluted argument involving matrix permutations. However, it dismissed all cases for $n\geq 4$ in a single paragraph by invoking Ryser's theorem from 1950. While bibliographical records confirm that Herbert Ryser published several papers on $(0,1)$-matrices during that period, the evaluators found that none of Ryser's theorems apply to the specific constraints of this problem. This application of irrelevant theory constitutes a clear ''hallucination'' or ''bluff''. Such passages serve as a distinct fingerprint, facilitating the differentiation between human-authored work and machine-generated content. For the sake of completeness, it should be noted that the lengthy solution provided by \flash was equally unsuccessful; cumulative errors in its extensive transformations led the model to the erroneous final conclusion that $n=2$ is the sole solution.

This task illustrates that the exhaustive, narrative style favored by Gemini models can be a double-edged sword. While these models do not shy away from detail, such verbosity increases the likelihood of cardinal errors or, more concerningly, the fabrication of exotic or entirely false mathematical facts. Conversely, the more concise models, by avoiding granular detail, frequently deliver incomplete solutions. In the context of this specific problem, no model achieved a golden mean between these two extremes.

Furthermore, it is striking to observe the occurrence of elementary errors in basic transitions, the type of mistakes one would not expect from a student participating in a rigorous competition, especially one who frequently demonstrates the ability to solve highly advanced problems.

\begin{center}
\begin{tcolorbox}[colback=white,colframe=black,boxrule=0.4pt,
arc=4mm,enhanced,width=0.9\textwidth,sharp corners=south]
\textbf{\textit{Set 4, Problem B5}}

\begin{tcolorbox}[colback=lightgray,colframe=black,arc=1mm,boxrule=0.2pt]

Let $V$ be a vector space over the field of rational numbers $Q$. Find all functions $f : V \to V$ satisfying the functional equation:
$$
f(f(z_1)+z_2)=z_1+f(z_2).
$$

\end{tcolorbox}
\url{https://github.com/google-deepmind/eval_hub/tree/master/eval_hub/putnam_like/putnam_like/Set_4/B5}

\end{tcolorbox}
\end{center}

A recurring issue in this task was the models' failure to comprehend the specific nature of the expected output. Both \rone and \pro concluded their derivations by stating that the sought functions are $\mathbb{Q}$-linear involutions. However, this response is insufficient; in the context of functional equations, the prompt ''Find all functions'' necessitates an explicit construction. Furthermore, \rone failed to verify whether these involutions actually satisfy the original equation, effectively proving only the necessary condition.

The \sonnet model correctly identified the necessary condition and noted that these involutions could take various forms, yet it failed to elaborate on this point. The \ofour model performed slightly better by suggesting a decomposition of the vector space $V$ into a sum of two subspaces and proposing a corresponding formula; nonetheless, it failed to rigorously demonstrate that this encompasses all possible involutions.

A complete logical trajectory was presented only by \othree and \flash. However, \othree committed a fundamental logical error, a circular implication (petitio principii), which critically undermined the validity of its proof and significantly lowered its score.

A key step in the reasoning for this problem involved demonstrating $\mathbb{Q}$-linearity by leveraging the function's additivity. This is a standard result rooted in the analysis of the classical Cauchy functional equation. Indeed, the authors of the problem treated this as a well-established fact, omitting its proof in the official solution. Interestingly, the models adopted two distinct approaches here: only \othree and \flash treated this property as a known lemma, whereas the others provided at least a skeletal proof. Even \rone, whose solutions are characteristically laconic, chose to present a sketch of the proof for this step.

\section*{Code and Data Availability and Reproducibility}

This section outlines the availability of code and data used in our analysis. The dataset is publicly available at \url{https://github.com/google-deepmind/eval_hub/tree/master/eval_hub/putnam_like}. All data is available under the Apache License 2.0.

\section*{Acknowledgements}
We would like to thank all the collaborators in the Putnam-like project, especially Henryk Michalewski, who invited us to contribute to this dataset.

\bibliographystyle{abbrv}
\bibliography{references}

\end{document}